\documentclass[11pt]{article}

\usepackage[preprint]{acl}

\usepackage{times}
\usepackage{latexsym}
\usepackage[T1]{fontenc}
\usepackage[utf8]{inputenc}
\usepackage{microtype}
\usepackage{inconsolata}

\usepackage{url}
\usepackage{graphicx}
\usepackage{booktabs}
\usepackage{tabularx}
\usepackage{tcolorbox}
\usepackage{pgfplots}
\pgfplotsset{compat=1.18}
\usepackage{multirow}
\usepackage{textcomp}
\usepackage{wrapfig}
\usepackage{pifont}
\usepackage{siunitx}
\usepackage{arydshln}
\usepackage{amsmath}
\usepackage{amssymb}
\usepackage{mathtools}
\usepackage{amsthm}

\newtheorem{theorem}{Theorem}

\newcommand{\uparrowcell}{\textcolor{green!50!black}{\ensuremath{\uparrow}}}
\newcommand{\downarrowcell}{\textcolor{red!85!black}{\ensuremath{\downarrow}}}

\title{BOW: Training Language Models to Reason Over Plausible Next Words}

\author{
Ming Shen\thanks{Equal contribution.},
Zhikun Xu\footnotemark[1],
Jacob Dineen, Xiao Ye, Ben Zhou\\[0.2em]
{Arizona State University}
}

\begin{document}
\maketitle

\begin{abstract}

Next-word prediction (NWP) trains language models against a single observed continuation, even though many contexts admit multiple plausible next words. Recent RL-based next-word reasoning methods make this tension explicit: they reward a model for producing a rationale that supports one context-conditioned continuation, which can turn a pre-existing preference into a confident, self-justifying trajectory. We introduce BOW, an RL framework that instead trains models to produce self-contained, neutral, and comprehensive descriptions of the plausible next-word space. The policy generates a next-word reasoning trajectory from the full context, but a frozen scorer computes the core reward from that trajectory alone, without separately receiving the context. BOW-Reg adds a lightweight breadth regularizer to this core reward to discourage premature collapse. On two model backbones, BOW remains competitive with the original models and often outperforms trained baselines across ten general reasoning benchmarks. On benchmarks testing ambiguous references and word meanings, BOW-Reg achieves the highest SharedRef correctness and the lowest HoWN-Simple single-sense collapse on both backbones. Human evaluation further shows that BOW-Reg produces broader next-word reasoning trajectories, while direct next-word-prediction evaluation shows that these trajectories remain predictive.\footnote{Code available at \url{https://github.com/ARC-ASU/BOW}.}

\end{abstract}
\section{Introduction}

Large language models (LLMs) are primarily trained with next-word prediction (NWP) during pretraining and post-training \citep{devlin-etal-2019-bert, JMLR:v21:20-074, NEURIPS2020_1457c0d6, Achiam2023GPT4TR}. Yet natural language is inherently underdetermined: a context often admits multiple plausible next words, whereas NWP rewards only the single continuation observed in the data. This mismatch can contribute to memorization-related bias \citep{lee-etal-2022-deduplicating, hans2024be, carlini2023quantifying}. This tension becomes more consequential in recent RL-based next-word reasoning methods such as Reinforcement Pre-Training (RPT) \citep{Dong2025ReinforcementP} and Reinforcement as a Pretraining Objective (RLP) \citep{hatamizadeh2026rlp}. Although effective, these methods remain answer-centric: they reward reasoning trajectories that support one context-conditioned continuation under the full input context. When the base model already prefers one plausible next word over others, such training can reinforce that skew\footnote{Here and throughout the paper, we use bias broadly to mean a pre-existing skew in the model's next-word distribution toward one plausible continuation, including corpus-frequency and lexical-shortcut effects; it is not restricted to demographic or social bias.} by turning a mild preference into a confident, self-justifying rationale.

\begin{figure*}[!t]
\centering
    \includegraphics[width=0.95\linewidth]{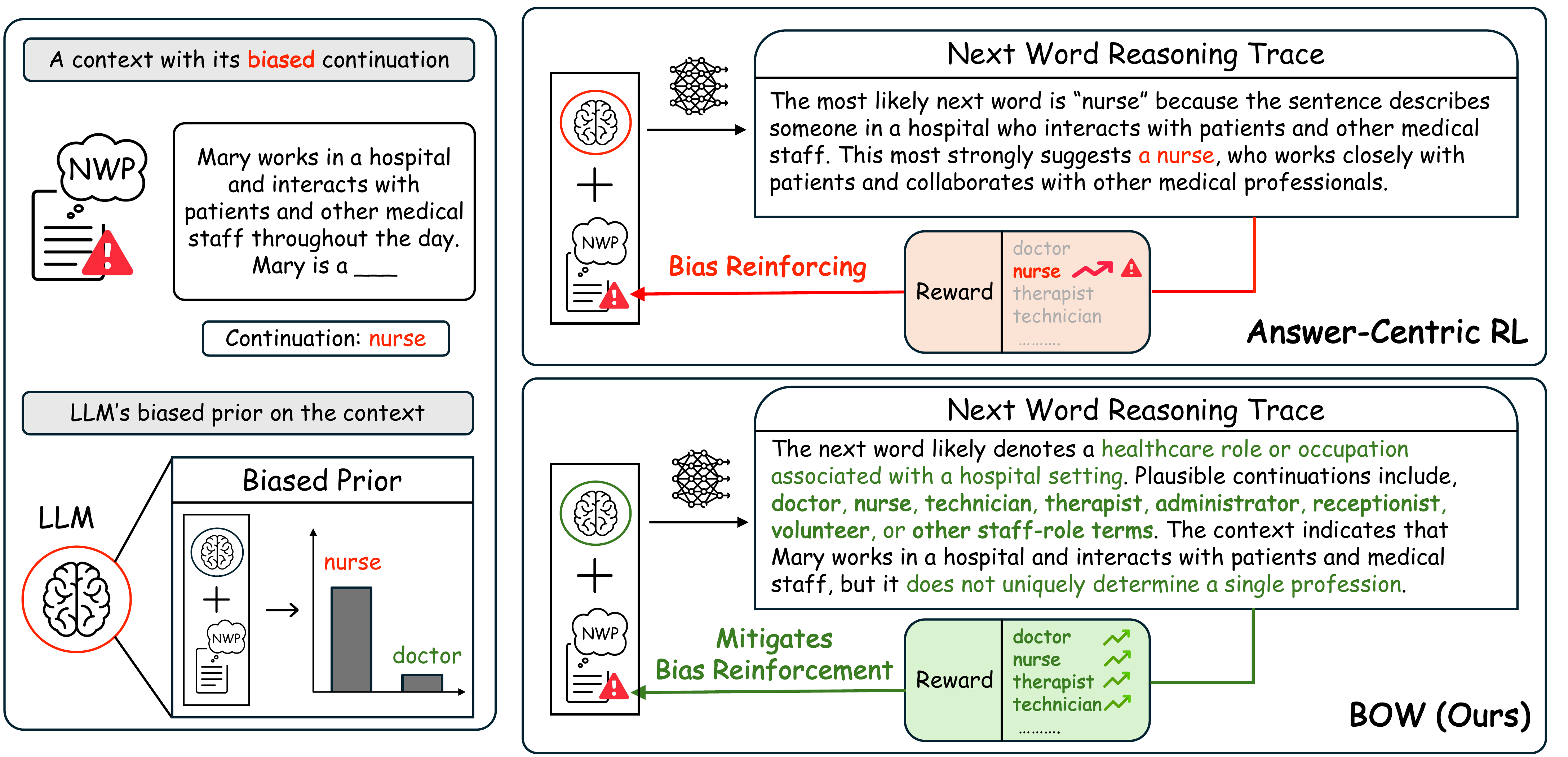}
    \caption{\small In an underdetermined context, the vanilla Qwen-based model assigns a higher probability to \textit{nurse} than to \textit{doctor}. Answer-centric RL rewards a trajectory that treats this favored continuation as uniquely implied, while BOW trains a neutral description of the broader plausible next-word space. Probabilities and trajectories are actual Qwen-based outputs.}
    \label{fig:bow_overview}
\end{figure*}

Fig.~\ref{fig:bow_overview} gives an intuitive example. Consider the context \textit{Mary works in a hospital and interacts with patients and other medical staff throughout the day. Mary is a}. The observed next word is \textit{nurse}, which the base model already favors over \textit{doctor}. The word \textit{nurse} is a valid continuation, but the context also supports other healthcare roles. An answer-centric objective can turn the model's prior preference for \textit{nurse} into a rationale that presents it as uniquely implied. The same failure can arise whenever several continuations remain plausible but the model already favors one of them. We refer to this as a bias-reinforcing mode of RL-based next-word reasoning.

This failure is especially likely because the policy is conditioned on a context that already induces a skewed next-token distribution, while the reward is tied to the observed continuation. A trajectory that turns the favored word into a confident explanation can therefore be reinforced even when the context still permits alternatives. Our goal is not to suppress the observed word, but to prevent its supervision from forcing the reasoning trajectory to treat one plausible continuation as uniquely licensed.
Prior work has shown that reducing inductive bias can generally improve LLM reasoning capabilities \citep{wang-etal-2023-causal, Zhang_Zhang_Wu_He_Zhou_2025, wang2025eliminating}. We hypothesize that removing this bias-reinforcing mode improves the model's general reasoning.
To avoid this mode, we train reasoning that preserves uncertainty over the plausible continuation space rather than collapsing immediately to a single answer.

We therefore propose BOW (Bottlenecked Next-Word Exploration).
BOW changes the target of the reasoning trajectory: instead of justifying a single favored continuation, the policy produces a self-contained, neutral, and comprehensive description of the plausible next-word space.
To make the trajectory responsible for the core reward, we remove the scorer’s direct access to the original context in the core scoring pass. The policy still conditions on the context, and a self-contained trajectory may restate the contextual constraints needed for prediction. The scorer, however, computes the core reward from the trajectory alone, without receiving context as a separate input. Context can therefore affect this reward only through information carried by the trajectory; there is no separate context-to-scorer path that bypasses the trajectory. App.~\ref{appendix:math_bottleneck} formalizes this distinction.
The core reward is the scorer's probability assigned to the gold next token, so partially informative trajectories can receive graded credit. This better matches linguistic uncertainty than hard exact-match supervision.
We prompt the model to follow this trajectory form rather than directly name a single answer. Prompting alone may not ensure this behavior, so BOW-Reg adds a lightweight breadth regularizer that discourages premature collapse to narrow answer naming. We treat BOW-Reg as the full method and report BOW-Core as an analysis variant that uses the core reward without the breadth regularizer.

Empirically, we adapt Qwen2.5-7B-Instruct and LLaMA3.1-8B-Instruct. 
Across ten general reasoning benchmarks, BOW remains competitive with the original instruction models and often outperforms both trained continual pretraining and RL baselines. 
Additional NWP-style adaptation on raw text can degrade instruction following, so matching or improving the original models indicates that BOW preserves broad capability. Targeted ambiguity and human evaluations test the collapse-mitigation claim directly.
To evaluate the targeted failure mode, we further test on SharedRef \citep{ellinger2025itdepends} and HoWN-Simple \citep{ellinger2025hown}, which assess referential ambiguity and lexical sense collapse. BOW-Reg is strongest on SharedRef and achieves the lowest HoWN-Simple single-sense collapse on both backbones. Human evaluation further shows that BOW-Reg elicits broader and more comprehensive next-word reasoning trajectories, whereas BOW-Core is more willing to verbalize a specific answer. 

Our contributions are: (i) characterizing answer-centric collapse as a limitation of next-word RL, (ii) proposing a core reward computed from the generated trajectory alone, together with a breadth regularizer, and (iii) showing across general reasoning, targeted ambiguity, intrinsic NWP, and human evaluation that this training principle improves reasoning quality while reducing premature collapse.

\section{Related Work}

Recent work has made intermediate reasoning explicit during language modeling. ToW \citep{xu-etal-2025-tow} and Rationalyst \citep{jiang-etal-2025-rationalyst} add supervised intermediate reasoning, while STaR \citep{zelikman2022star} and Quiet-STaR \citep{zelikman2024quietstar} use self-generated reasoning to improve prediction. These methods establish the value of reasoning before output, but their objectives do not directly train intermediate reasoning to preserve uncertainty over the next-word candidate space. BOW instead uses RL to train trajectories that emphasize contextual constraints and plausible continuations rather than solely supporting one realized word.

Closest to our setting, recent RL-based next-word reasoning methods such as RPT \citep{Dong2025ReinforcementP} and RLP \citep{hatamizadeh2026rlp} show that RL can be an effective post-training strategy for language modeling, but they remain answer-centric, rewarding trajectories for supporting a single favored continuation under the full context. More broadly, language models often exploit shortcuts, spurious correlations, or plausible-looking rationales rather than robust inference \citep{li-etal-2024-deceptive, Zhou2024ConceptualAU, Yu2024ReasonAgainUE}. We study this failure mode in next-word RL, where answer-centric training can turn an existing preference among plausible continuations into a more confident, self-justifying trajectory. BOW instead trains trajectories to describe the plausible next-word space rather than solely justify one realized continuation.

\section{Method}

We now formalize BOW. Let $C$ denote the input context and $w^\ast$ the supervised next token, i.e., the first token of the ground-truth next word; we use ``next word'' informally throughout. A policy language model $\pi_\theta$ first generates an intermediate next-word reasoning trajectory $\tau \sim \pi_\theta(\cdot \mid C)$. The policy conditions on the full context $C$ but never observes $w^\ast$; supervision arrives only through a scalar reward assigned after $\tau$ is produced.

\subsection{Reasoning Goal and Setup}

The goal of $\tau$ is not to name one answer as early as possible. Instead, it should reason over the plausible next-word space in a way that is informative enough to support prediction, yet broad enough to avoid premature answer commitment. This distinguishes BOW from recent answer-centric next-word RL: we train the model to reason over the continuation space rather than justify one favored continuation.

We ask each trajectory to satisfy three properties. Self-contained: it restates the contextual constraints needed for prediction rather than leaving them implicit in $C$. Neutral: it does not present one favored continuation as uniquely implied unless the candidate space is genuinely small. Comprehensive: it captures the main factors shaping the next word without exhaustively listing an open-ended candidate set. Explicit enumeration is allowed only when the plausible set is small and concrete.

Note that self-contained does not mean context-free. Because $\tau$ is generated from $C$, it may restate or summarize relevant parts of $C$. The bottleneck concerns the input to the core scoring pass, not whether $\tau$ repeats words from $C$.
The prompt specifies this trajectory form, while the learning objective determines which trajectories are reinforced. Fig.~\ref{fig:policy_prompt} gives the exact policy prompt.

\subsection{BOW-Core: Core Reward from the Trajectory Alone}

We place the bottleneck in \emph{credit assignment}: the policy conditions on the full context $C$, while the core scorer computes reward from $\tau$ alone. Any context-derived evidence used for this reward must first be represented in the trajectory.

We use a frozen scorer $J_\phi$. Given a reasoning trajectory $\tau$, the scorer is prompted to predict the immediate next token from $\tau$ alone, yielding a next-token distribution $P_{J_\phi}(\cdot \mid \tau)$. Here, $J_\phi$ is not a separately trained judge or reward model; it is a frozen off-the-shelf LM, and the gold next token is the only task-specific supervision used to compute the core reward. The core reward is
\begin{equation}
r_{\mathrm{core}}(\tau) := P_{J_\phi}(w^\ast \mid \tau),
\label{eq:core}
\end{equation}
where $w^\ast$ is the supervised next token. Because $r_{\mathrm{core}}$ is computed from $P_{J_\phi}(w^\ast \mid \tau)$, contextual evidence can affect gold-token credit only after the policy writes it into $\tau$. A scorer that separately received $C$ in the core scoring pass could assign high probability to $w^\ast$ while ignoring the trajectory, allowing the reward to bypass $\tau$. BOW-Core removes this route. This restriction does not prohibit overlap between $C$ and $\tau$; it makes the trajectory the required carrier of context-derived evidence used by the core scorer.
We call this objective BOW-Core, which lets us evaluate the effect of computing the core reward from the generated trajectory alone. App.~\ref{appendix:math_bottleneck} formalizes its reward-path bottleneck. Fig.~\ref{fig:scorer_justification} illustrates how candidate next words encoded in $\tau$ can appear among the scorer's high-ranked tokens, and Fig.~\ref{fig:judge_prompt} gives the scorer prompt.


BOW-Core also yields a soft, distributional reward rather than a binary one. A trajectory that substantially narrows the plausible continuation space, even without identifying the exact gold next word, can still receive graded credit. This better matches linguistic underdetermination and provides a denser learning signal than exact-match rewards. App.~\ref{appendix:math_soft_vs_binary} further motivates the partial-credit advantage of the soft reward.

\subsection{BOW-Reg: The Full Objective with Lightweight Breadth Regularization}

BOW-Core encourages the intended trajectory form, but it does not fully prevent collapse to narrow, answer-like trajectories. We sometimes observe trajectories that enumerate only a few candidate words or wrap the current favorite answer in slightly longer prose during training. This can be appropriate when the candidate set is genuinely small, but not when the context still supports a broader range of continuations. BOW-Reg therefore adds a lightweight breadth regularizer to discourage this behavior.

Using the same frozen scorer, we take $P_{J_\phi}(\cdot \mid C)$ as a reference distribution and define
\begin{equation}
\begin{split}
R_{\mathrm{reg}}(\tau) :=
&\ r_{\mathrm{core}}(\tau)
- \alpha
\left\|
P_{J_\phi}(\cdot \mid \tau)
\right. \\
&\left.
- P_{J_\phi}(\cdot \mid C)
\right\|_1 .
\end{split}
\label{eq:reg}
\end{equation}
where $\alpha \ge 0$ controls the strength of the regularizer.
The regularizer shapes rather than replaces the core objective; by itself, it does not guarantee unbiased behavior. It discourages trajectories whose induced next-token distribution drifts too far from the reference; in practice, such drift often corresponds to premature answer naming. We interpret Eq.~\ref{eq:reg} as a \emph{breadth regularizer}: the prompt specifies the desired trajectory form, while the regularizer helps preserve it when optimization under the core reward becomes too narrow.
As a result, BOW-Core uses only the gold-token term $P_{J_\phi}(w^\ast \mid \tau)$, whereas BOW-Reg adds the breadth regularizer in Eq.~\ref{eq:reg}.

\subsection{Policy Optimization}

Given a training pair $(C, w^\ast)$, the policy samples a group of reasoning trajectories $\{\tau_1, \ldots, \tau_N\}$ from $\pi_\theta(\cdot \mid C)$. For each $\tau_i$, we compute either $r_{\mathrm{core}}(\tau_i)$ for BOW-Core or $R_{\mathrm{reg}}(\tau_i)$ for BOW-Reg. The policy is then trained to maximize expected reward.
In our experiments, we optimize this objective with Group Relative Policy Optimization (GRPO) \citep{Shao2024DeepSeekMathPT}. For each context, rewards are normalized within the sampled group to obtain relative advantages, and the policy is updated with PPO-style steps \citep{Schulman2017ProximalPO}. Group normalization is useful because the absolute reward scale can vary across contexts with different levels of next-word uncertainty. The method is not tied to GRPO; standard policy-gradient methods such as PPO \citep{Schulman2017ProximalPO} or REINFORCE \citep{Williams2004SimpleSG} can also be used.
\section{Experimental Setup}

We evaluate BOW on four task-facing axes: general zero-shot reasoning, targeted ambiguity and collapse, human judgments of candidate-space reasoning, and intrinsic next-word prediction. We additionally test robustness to scorer family, scorer size, and reward sharpness.

\subsection{Training Data}

We construct the RL training set to emphasize reasoning-critical next-word prediction: contexts where multiple next words are plausible but deeper contextual reasoning is needed to narrow the candidate space. Starting from long-form narrative prose drawn from prior work \citep{del-fishel-2023-true}, we apply a reasoning-centric filter that removes locally trivial continuations, such as function-word completions and highly formulaic collocations. This yields about 45K context--next-word pairs, where accurate prediction typically requires tracking entities, events, and implicit world dynamics across multiple sentences. Although the source texts are narrative, the selection criterion is task-driven rather than genre-specific: we retain only positions where the next word requires non-trivial inference from the preceding text.
App.~\ref{appendix:training_data_filtering} describes the filter, and App.~\ref{sec:data_filtering} reports an ablation against random filtering. The filter provides modest but consistent gains by concentrating updates on contexts where next-word reasoning is needed.

\subsection{Models and Optimization}

We use Qwen2.5-7B-Instruct \citep{Yang2024Qwen25TR} and LLaMA3.1-8B-Instruct \citep{Dubey2024TheL3} as policy models. For both model families, we train for one epoch with a total batch size of 1024, a mini-batch size of 256, and a rollout size of 5. We disable the KL term used in standard GRPO following recent works \citep{liu2025understanding, yu2025dapo}.

We use LLaMA3.1-8B-Instruct as the default frozen scorer $J_\phi$ because its trajectory-conditioned distributions are less sharply concentrated. Sec.~\ref{sec:scorer_robustness} analyzes this choice and tests robustness to scorer family, scorer size, and reward sharpness.
To obtain $P_{J_\phi}(\cdot \mid \tau)$ in the core reward, we use temperature $T=5$ and retain the top 100 tokens of the scorer distribution. If the first token of the gold next word does not appear in this set, the trajectory receives reward 0. For BOW-Reg, we use the same top-100 truncation for both distributions in Eq.~\ref{eq:reg} and set the regularization strength to $\alpha = 0.1$.\footnote{Set once from the observed training-time magnitudes of the two terms, without tuning on evaluation performance, so the regularizer remained influential without dominating the core reward.}
Additional implementation details, including training-time compute, are reported in App.~\ref{appendix:more_optim_details}.

\subsection{Baselines}

We compare BOW against two continual-pretraining baselines, two answer-centric RL baselines, and the original instruction-tuned models for reference. For fair comparison, all trainable baselines use the same filtered supervision positions as BOW. The continual-pretraining baselines are SLM \citep{lin2024not}, which applies selective language modeling to the same context--next-word pairs, and ToW \citep{xu-etal-2025-tow}, which augments those positions with distilled word-level thoughts from a stronger annotator model before standard causal language modeling.

The answer-centric RL baselines are BR (Binary-Reward RL), an RPT-style \citep{Dong2025ReinforcementP} objective that reasons and then commits to a single next-word prediction with exact-match reward on the first token, and IGR (Information-Gain RL), a simplified RLP-style \citep{hatamizadeh2026rlp} objective that uses a frozen reference scorer to reward the difference between reasoned and baseline log-evidence. We also report the original Qwen2.5-7B-Instruct and LLaMA3.1-8B-Instruct models without additional training, since further NWP-style adaptation can affect instruction-following behavior. 

\subsection{Evaluation Protocol}

\begin{table*}[t]
\centering
\resizebox{1\linewidth}{!}{%
\setlength{\tabcolsep}{6pt}
\small
\begin{tabular}{lcccccccccc}
\toprule
 & \textbf{CSQA} & \textbf{PIQA}  & \textbf{SQA} & \textbf{TQA} & \textbf{ARC-c} & \textbf{WG} & \textbf{BBH} & \textbf{MMLU} & \textbf{MMLU-p} & \textbf{GPQA}  \\
\midrule \\[-13pt]
\multicolumn{11}{c}{\footnotesize\texttt{\textit{Qwen2.5-7B-Instruct}}} \\[-2.5pt]
\midrule
Vanilla & 80.51 & 87.21 & 64.21 & 64.65 & 92.06 & 64.48 & 77.76 & 75.28 & 58.53 & 31.92 \\
\cdashline{1-11} \\[-8pt]
 SLM & 75.51 & 84.55 & 62.99 & 53.98 & 88.05 & 59.43 & 54.92 & 69.53 & 48.11 & 29.24 \\
 ToW & 81.16 & 88.41 & \textbf{67.69} & 57.16 & \textbf{93.00} & 66.14 & 69.21 & 69.52 & 55.16 & 32.14 \\
 BR & 77.64 & 87.05 & 60.53 & 61.08 & 89.76 & 62.59 & 76.82 & 70.82 & 46.17 & 31.25 \\
 IGR & 78.71 & 87.27 & 63.39 & 65.73 & 92.41 & 68.43 & \textbf{78.01} & 75.60 & 56.94 & 29.69 \\
 \textbf{BOW-Core} & 81.49 & 87.21 & 64.83 & 63.28 & 92.75 & 64.56 & 76.78 & 75.69 & \textbf{58.22} & 32.37 \\
 \textbf{BOW-Reg} & \textbf{81.90} & \textbf{88.68} & 63.19 & \textbf{66.83} & 92.32 & \textbf{70.96} & 77.21 & \textbf{76.51} & 57.51 & \textbf{33.26} \\

\midrule \\[-13pt]
\multicolumn{11}{c}{\footnotesize\texttt{\textit{LLaMA3.1-8B-Instruct}}} \\[-2.5pt]
\midrule
 Vanilla & 77.07 & 85.15 & 69.33 & 60.47 & 89.25 & 59.59 & 74.71 & 72.90 & 50.50 & 32.14 \\
\cdashline{1-11} \\[-8pt]
 SLM & 36.12 & 67.85 & 53.78 & 9.55 & 45.82 & 52.17 & 22.01 & 39.32 & 12.80 & 24.55 \\
 ToW & 22.28 & 60.07 & 46.83 & 28.64 & 34.30 & 45.07 & 22.71 & 43.77 & 21.16 & 25.00 \\
 BR & 76.66 & 83.73 & 66.46 & 53.12 & 84.47 & 57.14 & 65.89 & 66.33 & 38.56 & 22.32 \\
 IGR & 77.81 & 86.62 & 67.89 & 58.63 & 90.02 & 60.06 & 73.98 & 73.27 & 49.78 & 31.03 \\
 \textbf{BOW-Core} & \textbf{78.71} & \textbf{87.11} & \textbf{68.92} & 58.75 & \textbf{90.36} & \textbf{61.96} & 74.10 & \textbf{74.21} & \textbf{53.13} & \textbf{35.04} \\
 \textbf{BOW-Reg} & 77.97 & 86.67 & 67.28 & \textbf{60.22} & 89.16 & 59.75 & \textbf{74.56} & 73.59 & 52.71 & 32.14 \\

\bottomrule
\end{tabular}
}
\caption{\small General reasoning evaluation of BOW-Core, BOW-Reg, and baselines. Bold indicates the best result among trained methods. Vanilla instruction models are shown as reference points and are not included in comparisons among trained methods. We use SC@10 (self-consistency) for all benchmarks. TQA = TruthfulQA, SQA = StrategyQA, ARC-c = ARC-Challenge, WG = WinoGrande, and MMLU-p = MMLU-Pro.}
\label{tab:general_evaluation}
\end{table*}

For general reasoning evaluation, we use ten benchmarks spanning commonsense, world knowledge, and multi-hop reasoning: CSQA \citep{talmor-etal-2019-commonsenseqa}, PIQA \citep{bras_Gao_Choi_2020}, StrategyQA \citep{geva-etal-2021-aristotle}, TruthfulQA \citep{lin-etal-2022-truthfulqa}, ARC-Challenge \citep{Clark2018ThinkYH}, WinoGrande \citep{Bras_Bhagavatula_Choi_2020}, BBH \citep{suzgun-etal-2023-challenging}, MMLU \citep{hendrycks2021measuring}, MMLU-Pro \citep{wang2024mmlupro}, and GPQA \citep{rein2024gpqa}. We use each benchmark in its original multiple-choice QA format. We perform zero-shot inference by prompting the policy to think step by step before outputting the final answer letter. We use Math-Verify\footnote{\url{https://github.com/huggingface/Math-Verify}} to extract the answer letter. Fig.~\ref{fig:mcqa_inference_prompt} gives the inference prompt.

For targeted ambiguity evaluation, we evaluate the English SharedRef split under the Normal prompt \citep{ellinger2025itdepends} and the English HoWN-Simple setting \citep{ellinger2025hown}. SharedRef reports official judge-based Correctness under referential ambiguity. HoWN-Simple evaluates homonym definitions in response to a simplified-language request; we report the official Covered and single-sense Collapse. These metrics assess whether a model preserves plausible referents or senses rather than prematurely committing to a single one. App.~\ref{appendix:ambiguity_details} provides additional dataset and evaluation details.

For intrinsic next-word prediction evaluation, we convert five benchmarks into multiple-choice next-word prediction tasks while preserving the original reasoning requirement: CSQA, PIQA, StrategyQA, ARC-Challenge, and WinoGrande. The policy first generates a next-word reasoning trajectory; the frozen scorer then scores the candidate next words using the same trajectory-conditioned prompt as in training, and the highest-probability candidate is taken as the prediction. Following prior work, we do not report intrinsic NWP for IGR. For BOW, we report only BOW-Reg because this evaluation assesses the full BOW-Reg objective; BOW-Core is an analysis variant that omits the breadth regularizer. Tab.~\ref{tab:transformation_examples} shows two transformed examples, and Apps.~\ref{appendix:data_transformation} and~\ref{appendix:more_baseline_details} provide additional transformation and baseline details.

For the general reasoning and intrinsic next-word prediction evaluations, we use vLLM \citep{10.1145/3600006.3613165} and report self-consistency \citep{wang2023selfconsistency} with 10 samples (SC@10). We sample 10 reasoning trajectories at temperature 0.8 and take a majority vote over the resulting predictions. Neither SharedRef nor HoWN-Simple is a multiple-choice evaluation; for both, we generate one response per instance and apply the official evaluation schema.

\section{Results}


\subsection{General Zero-Shot Reasoning}
\label{sec:main-results}

Tab.~\ref{tab:general_evaluation} reports SC@10 on ten reasoning benchmarks. Across the two model families, a BOW variant is the strongest trained method on 17 of 20 model--benchmark pairs. These results show that training over the plausible next-word space preserves general reasoning performance and often improves over trained baselines, even though BOW is not optimized for downstream QA.

On Qwen, BOW-Reg achieves the best average score among trained methods, with BOW-Core close behind; together the two variants are strongest on 7 of the 10 benchmarks. On LLaMA, BOW-Core achieves the best average score, and the two variants are strongest on all of 10 benchmarks. In both model families, BOW remains competitive with and often slightly above the original instruction-tuned model despite the risk that additional next-word adaptation can weaken instruction following.

The continual-pretraining baselines are less stable across model families. On Qwen, ToW remains competitive on a few tasks but trails BOW on average despite using distilled supervision from a stronger annotator model. On LLaMA, SLM and ToW degrade sharply due to abstentions and degenerate generations, whereas both BOW variants remain stable. 
BOW-Core shows that the core reward provides a strong training signal even without breadth regularization. BOW-Reg preserves this performance while more directly targeting the ambiguity-collapse behavior evaluated next.

\subsection{Targeted Ambiguity and Premature-Collapse Evaluation}
\label{sec:ambiguity_eval}

\begin{table}[t]
\centering
\resizebox{0.95\columnwidth}{!}{%
\begin{tabular}{@{}lccc@{}}
\toprule
 & \multicolumn{1}{c}{\textbf{SharedRef}}
 & \multicolumn{2}{c}{\textbf{HoWN-Simple}} \\
\cmidrule(lr){2-2}\cmidrule(lr){3-4}
 & \textbf{Correctness} $\uparrow$
 & \textbf{Covered} $\uparrow$
 & \textbf{Collapse} $\downarrow$ \\
\midrule \\[-18pt]
\multicolumn{4}{c}{\footnotesize\texttt{\textit{Qwen2.5-7B-Instruct}}} \\[-3pt]
\midrule
Vanilla & 36.12 & 48.64 & 48.15 \\
\cdashline{1-4} \\[-10pt]
SLM     & 36.12 & 47.48 & 41.10 \\
ToW     & 66.52 & 48.65 & 37.68 \\
BR      & 56.39 & \textbf{52.20} & 48.47 \\
IGR     & 15.42 & 45.77 & 61.11 \\
\textbf{BOW-Core} & 50.66 & 51.19 & 44.44 \\
\textbf{BOW-Reg}  & \textbf{78.85} & 50.84 & \textbf{31.90} \\
\midrule \\[-18pt]
\multicolumn{4}{c}{\footnotesize\texttt{\textit{LLaMA3.1-8B-Instruct}}} \\[-3pt]
\midrule
Vanilla & 14.10 & 59.20 & 28.22 \\
\cdashline{1-4} \\[-10pt]
SLM     & 28.40 & 13.20 & 85.37 \\
ToW     & 15.86 & 49.75 & 36.20 \\
BR      & 13.66 & 56.38 & 25.00 \\
IGR     & 15.86 & 55.67 & 28.22 \\
\textbf{BOW-Core} & 21.15 & \textbf{65.86} & 8.59 \\
\textbf{BOW-Reg}  & \textbf{29.07} & 63.91 & \textbf{7.36} \\
\bottomrule
\end{tabular}}
\caption{\small Targeted ambiguity evaluation on English SharedRef (Normal) and HoWN-Simple. All scores are percentages. Bold indicates the best result among trained methods.}
\label{tab:ambiguity_eval}
\end{table}

Tab.~\ref{tab:ambiguity_eval} reports targeted ambiguity results. On SharedRef, BOW-Reg reaches 78.85 with Qwen, outperforming the strongest non-BOW baseline, ToW at 66.52, as well as BR at 56.39 and BOW-Core at 50.66. With LLaMA, BOW-Reg reaches 29.07, slightly above SLM at 28.40 and above BOW-Core at 21.15, IGR and ToW at 15.86, and BR at 13.66.

On HoWN-Simple, BOW-Reg achieves the lowest single-sense collapse on both backbones: 31.90 with Qwen and 7.36 with LLaMA. It does not uniformly maximize coverage: BR is highest with Qwen, while BOW-Core is highest with LLaMA. Even so, BOW-Reg remains competitive on coverage while most consistently avoiding single-sense collapse. The two HoWN metrics capture different aspects of behavior: Covered measures how many senses are mentioned, while Collapse records whether the response reduces the word to one sense. BOW-Reg does not simply maximize the number of senses mentioned; it most consistently controls the single-sense failure mode targeted by the method.

The non-BOW baselines are less consistent across settings. ToW is the strongest non-BOW method on Qwen SharedRef, while SLM is competitive on LLaMA SharedRef, but neither combines strong referential-ambiguity performance with low HoWN-Simple collapse across both backbones. SLM, for example, reaches 85.37 collapse with LLaMA despite its relatively strong SharedRef result. Among the evaluated methods, BOW-Reg is the only one that achieves both the highest SharedRef correctness and the lowest HoWN-Simple collapse for both model families. This consistency matters because the targeted failure mode spans both referential and lexical ambiguity, rather than performance on a single benchmark.

Together, the two benchmarks separate the roles of the variants. BOW-Core can preserve strong sense coverage, whereas BOW-Reg more strongly controls premature commitment. We therefore use BOW-Reg as the main method for the failure mode studied here, while BOW-Core serves as the analysis variant without breadth regularization.

\begin{figure}[t]
  \centering
  \definecolor{SoftBlue}{RGB}{114,147,203}
  \definecolor{SoftGreen}{RGB}{132,186,91}
  \definecolor{SoftOrange}{RGB}{225,151,76}
  \definecolor{SoftPurple}{RGB}{144,103,167}
  \scalebox{0.7}{%
  \begin{tikzpicture}
  \begin{axis}[
      height=5.5cm,
      ybar,
      x=2.9cm,
      enlarge x limits=0.45,
      clip=false,
      bar width=11pt,
      ylabel={Preference Rate (\%)},
      ylabel style={font=\large},
      xlabel style={font=\large},
      symbolic x coords={Comp., Next-Word Desc.},
      xtick=data,
      x tick label style={font=\small, rotate=0, anchor=north},
      y tick label style={font=\small},
      nodes near coords,
      nodes near coords style={font=\small, rotate=0, anchor=south},
      nodes near coords align={vertical},
      ymin=15,
      ymax=95,
      ymajorgrids=true,
      grid style={dashed, gray!30},
      legend style={
          at={(0.5,-0.18)},
          anchor=north,
          legend columns=5,
          font=\small,
          draw=none,
          /tikz/every even column/.append style={column sep=0.5em}
      },
      every axis plot/.append style={fill opacity=0.85},
  ]
  \addplot[fill=SoftBlue,   draw=black, line width=0.5pt] coordinates {(Comp.,48)(Next-Word Desc.,69)};
  \addplot[fill=SoftGreen,  draw=black, line width=0.5pt] coordinates {(Comp.,25)(Next-Word Desc.,67)};
  \addplot[fill=SoftOrange, draw=black, line width=0.5pt] coordinates {(Comp.,52)(Next-Word Desc.,81)};
  \addplot[fill=SoftPurple, draw=black, line width=0.5pt] coordinates {(Comp.,83)(Next-Word Desc.,41)};
  \legend{Vanilla, BR, BOW-Core, BOW-Reg}
  \end{axis}
  \end{tikzpicture}
  }
  \caption{\small Human-evaluation preference rates for comprehensive next-word reasoning and expected next-word description over 150 contexts. Ties were allowed, so rates need not sum to 100\%.}
  \label{fig:preference_annotation}
\end{figure}

\begin{table*}[t]
\centering
\resizebox{1\linewidth}{!}{%
\setlength{\tabcolsep}{6pt}
\small
\begin{tabular}{
l
*{11}{S[table-format=+2.2, table-space-text-post=\downarrowcell]}
}
\toprule
 & \multicolumn{1}{c}{\textbf{CSQA}}
 & \multicolumn{1}{c}{\textbf{PIQA}}
 & \multicolumn{1}{c}{\textbf{SQA}}
 & \multicolumn{1}{c}{\textbf{TQA}}
 & \multicolumn{1}{c}{\textbf{ARC-c}}
 & \multicolumn{1}{c}{\textbf{WG}}
 & \multicolumn{1}{c}{\textbf{BBH}}
 & \multicolumn{1}{c}{\textbf{MMLU}}
 & \multicolumn{1}{c}{\textbf{MMLU-p}}
 & \multicolumn{1}{c}{\textbf{GPQA}}
 & \multicolumn{1}{c}{\textbf{Avg.}} \\
\midrule

BOW-Core
 & 81.49 & 87.21 & 64.83 & 63.28 & 92.75 & 64.56
 & 76.78 & 75.69 & 58.22 & 32.37 & 69.72 \\

\midrule

\ding{172} $J_{\phi} \coloneqq \text{\texttt{Q2.5-7B-I}}$
 & -0.49\downarrowcell & -0.27\downarrowcell & +0.41\uparrowcell
 & +3.31\uparrowcell & -0.69\downarrowcell & +6.32\uparrowcell
 & +0.59\uparrowcell & +0.55\uparrowcell & -0.32\downarrowcell
 & -0.67\downarrowcell & +0.87\uparrowcell \\

\ding{173} $J_{\phi} \coloneqq \text{\texttt{L3.2-1B-I}}$
 & +0.16\uparrowcell & +0.66\uparrowcell & -0.82\downarrowcell
 & +0.86\uparrowcell & +0.34\uparrowcell & +4.03\uparrowcell
 & +0.43\uparrowcell & +0.37\uparrowcell & +0.02\uparrowcell
 & -3.35\downarrowcell & +0.27\uparrowcell \\

\ding{174} $T=2,\ \text{top-}50$
 & +1.07\uparrowcell & +1.69\uparrowcell & +0.81\uparrowcell
 & +0.49\uparrowcell & +0.42\uparrowcell & +6.47\uparrowcell
 & +0.59\uparrowcell & +0.61\uparrowcell & +0.76\uparrowcell
 & -0.90\downarrowcell & +1.20\uparrowcell \\

\bottomrule
\end{tabular}
}
\caption{\small Scorer robustness for BOW-Core with a Qwen2.5-7B-Instruct policy. The default configuration uses LLaMA3.1-8B-Instruct with $T=5$ and top-100 truncation. The BOW-Core row reports absolute scores; ablation rows report changes relative to this configuration. Q2.5-7B-I and L3.2-1B-I denote Qwen2.5-7B-Instruct and LLaMA3.2-1B-Instruct.}
\label{tab:scorer_robust}
\end{table*}

\subsection{Human Evaluation of Next-Word Reasoning}
\label{sec:human_results}

Benchmark accuracy does not reveal the form of the next-word reasoning trajectories. We therefore complement Tab.~\ref{tab:general_evaluation} and~\ref{tab:ambiguity_eval} with a human study of the trajectories themselves.

We curate 150 contexts where several immediate next words are superficially plausible but deeper contextual reasoning narrows the continuation space. We compare four Qwen-based models: the vanilla instruction model, BR, BOW-Core, and BOW-Reg. Annotators view anonymized, shuffled trajectories and judge which one best reasons over the plausible next-word space in a self-contained and comprehensive way, and which next-word description best matches their expected continuation. Ties are allowed. App.~\ref{appendix:more_human_study_details} provides the full curation and annotation protocol.

Fig.~\ref{fig:preference_annotation} provides behavioral evidence. The comprehensiveness dimension evaluates the broader trajectory form targeted by BOW. BOW-Reg receives an 83\% preference rate on this dimension, compared with 52\% for BOW-Core, 48\% for the vanilla model, and 25\% for BR. For expected next-word description, BOW-Core leads at 81\%, followed by the vanilla model at 69\% and BR at 67\%, while BOW-Reg receives 41\%. Because ties are allowed, the preference rates need not sum to 100.

This reversal reflects the two objectives. BOW-Core more readily states a likely continuation, which aligns with the direct answer an annotator expects. BOW-Reg preserves a broader candidate-space description and is therefore less likely to verbalize one answer immediately. BOW-Core's gain over BR on comprehensiveness also indicates that computing the core reward from the trajectory alone changes the learned reasoning behavior, even without the breadth regularizer. App.~\ref{appendix:more_human_study_details} provides representative examples from ordinary ambiguous, medical, and finance contexts.

\begin{table}[t]
  \centering
  \resizebox{0.95\columnwidth}{!}{%
  \begin{tabular}{@{}lccccc@{}}
    \toprule
     & \textbf{CSQA} & \textbf{PIQA} & \textbf{SQA} & \textbf{ARC-c} & \textbf{WG} \\
    \midrule \\[-18pt]
    \multicolumn{6}{c}{\footnotesize\texttt{\textit{Qwen2.5-7B-Instruct}}} \\[-3pt]
    \midrule
     Vanilla & 84.26 & 85.46 & 50.92 & 80.52 & 67.70 \\
     \cdashline{1-6} \\[-10pt]
     SLM     & 79.73 & 80.50 & 59.51 & 75.68 & 55.49 \\
     ToW     & 87.55 & 87.16 & 57.06 & 80.12 & 67.81 \\
     BR      & 84.00 & 86.89 & 63.80 & \textbf{83.24} & \textbf{74.92} \\
     \textbf{BOW-Reg} & \textbf{88.40} & \textbf{89.06} & \textbf{64.21} & 80.02 & 66.81 \\
    \midrule \\[-18pt]
    \multicolumn{6}{c}{\footnotesize\texttt{\textit{LLaMA3.1-8B-Instruct}}} \\[-3pt]
    \midrule
     Vanilla & 84.74 & 83.56 & 54.19 & 73.16 & 56.94 \\
     \cdashline{1-6} \\[-10pt]
     SLM     & 40.66 & 59.65 & 49.28 & 37.54 & 50.28 \\
     ToW     & 87.79 & 88.65 & 58.28 & 79.02 & 66.26 \\
     BR      & 82.30 & 88.86 & 57.06 & 78.00 & \textbf{70.14} \\
     \textbf{BOW-Reg} & \textbf{87.91} & \textbf{89.27} & \textbf{62.17} & \textbf{79.11} & 64.71 \\
    \bottomrule
  \end{tabular}
  }
  \caption{\small Intrinsic next-word recovery of BOW-Reg and various baselines. All scores are SC@10. Bold indicates the best result among trained methods.}
  \label{tab:nwp_evaluation}
\end{table}

\subsection{Intrinsic Next-Word Prediction}
\label{sec:intrinsic_evaluation}

Tab.~\ref{tab:nwp_evaluation} reports intrinsic next-word prediction on five benchmarks converted into multiple-choice NWP form. We evaluate only BOW-Reg because this setting assesses the full method.

BOW-Reg achieves the best result in 7 of the 10 model--benchmark settings. On LLaMA, it is the strongest method on 4 of the 5 tasks and also has the best average score. On Qwen, it leads on CSQA, PIQA, and StrategyQA, while trailing BR on ARC-Challenge and WinoGrande. A breadth-oriented objective could, in principle, make the trajectories too diffuse for accurate prediction. Here, candidate scores are computed from the generated trajectory rather than the original context, so the evaluation directly tests whether BOW-Reg's broader trajectories remain predictive representations. The results show that reducing early commitment does not remove task-relevant information: the trajectories still support accurate continuation recovery.

\subsection{Scorer Robustness}
\label{sec:scorer_robustness}

Because BOW uses a frozen scorer to assign the core reward, we test whether the resulting training signal depends on the scorer configuration. We run this ablation with the Qwen2.5-7B-Instruct policy and BOW-Core, which excludes the breadth regularizer so that the comparison focuses on the core reward. Our default scorer is LLaMA3.1-8B-Instruct. In the qualitative example in Fig.~\ref{fig:scorer_justification} (App.~\ref{appendix:math_soft_vs_binary}), both LLaMA3.1-8B-Instruct and Qwen2.5-7B-Instruct rank \textit{water} first, but Qwen concentrates nearly all probability mass on it (0.9995), whereas LLaMA remains less concentrated (0.7575) and assigns non-negligible mass to other candidates. We therefore use LLaMA as the default because this scoring behavior better matches the goal of rewarding neutral descriptions of the plausible next-word space.

Tab.~\ref{tab:scorer_robust} compares the default configuration with three variants: a Qwen2.5-7B-Instruct scorer, a smaller LLaMA3.2-1B-Instruct scorer, and a sharper reward distribution using $T=2$ and top-50 truncation instead of $T=5$ and top-100. The variants change average performance by $+0.87$, $+0.27$, and $+1.20$ points, respectively, although individual benchmarks vary. Thus, the core reward is robust on average to the tested changes in scorer family, model size, and reward sharpness; it does not rely on the default scorer configuration.
\section{Conclusion}
We introduced BOW, an RL framework that trains next-word reasoning trajectories to describe the plausible continuation space rather than justify a single observed continuation.
BOW-Core isolates the effect of computing the core reward from the generated trajectory alone, and BOW-Reg adds breadth regularization to reduce premature collapse while preserving general reasoning and intrinsic NWP performance.
Across both backbones, BOW-Reg achieves the highest SharedRef correctness and the lowest HoWN-Simple collapse; in the Qwen-based human study, it receives the highest comprehensiveness preference.
The scorer ablation further shows that the core reward is robust on average across the tested scorer configurations.
Together, these findings support training next-word reasoning to preserve plausible alternatives until the context justifies commitment.




\section*{Limitations}



BOW targets a specific failure mode in RL-based next-word reasoning: when several continuations remain plausible, answer-centric rewards can push the model to commit too early to one favored continuation and generate reasoning that supports it. BOW should therefore not be interpreted as a general-purpose method for maximizing performance on every downstream QA benchmark. We use general QA and intrinsic NWP primarily to verify that the intervention preserves broad capability, while SharedRef, HoWN-Simple, and the human study provide more direct tests of the premature-collapse behavior that BOW is designed to address.

Our experiments cover two 7--8B instruction backbones, English ambiguity benchmarks, a Qwen-based human study, and reasoning-heavy next-word positions selected by filtering. They test the proposed failure mode in a controlled setting but do not establish how BOW behaves in larger-scale pretraining, multilingual data, specialized domains, or longer-form generation. 

\bibliography{custom}

\appendix

\section{Reward-Path Bottleneck Analysis of the Core BOW Reward}
\label{appendix:math_bottleneck}

\paragraph{Motivation.}
A central goal of the core BOW reward is to make the generated trajectory $\tau$ responsible for credit assignment in next-word learning. The policy still conditions on $C$, so $\tau$ may restate or summarize information from $C$. In the core scoring pass, however, the scorer receives only $\tau$, without receiving $C$ as a separate input. This appendix formalizes why this design pressures $\tau$ to remain predictive of the next token, whereas joint scoring from $(C,\tau)$ can bypass the trajectory. The analysis concerns the core reward in Eq.~\ref{eq:core}, which BOW-Reg retains; the breadth term in Eq.~\ref{eq:reg} is outside its scope.

\subsection{Setup and Notation}

Let $C := x_{1:t}$ denote the context and let $Y := x^*_{t+1}$ denote the gold next token (or first token of the gold next word, consistent with our implementation). Let $\tau$ be the next-word reasoning trajectory sampled from the policy:
\[
\tau \sim \pi_\theta(\cdot \mid C).
\]
Since $\tau$ is generated from $C$ and does not influence the data-generating process of $Y$, we have the Markov structure
\[
\begin{gathered}
\tau \leftarrow C \rightarrow Y, \\
\text{equivalently} \qquad I(Y;\tau \mid C)=0.
\end{gathered}
\]

We first contrast two scoring schemes:

\paragraph{Core scoring from the trajectory alone (BOW-Core).}
A frozen scorer $J_\phi$ receives only $\tau$ and outputs a next-token distribution $P_{J_\phi}(\cdot \mid \tau)$. The core BOW reward is
\[
r_{\textsc{bow}}(\tau;Y) := P_{J_\phi}(Y \mid \tau).
\]

\paragraph{Joint context-and-trajectory scoring.}
A probabilistic scorer $\tilde J_\phi$ receives the original context $C$ in addition to $\tau$, and assigns reward via
\[
r_{\textsc{ctx}}(C,\tau;Y) := \log P_{\tilde J_\phi}(Y \mid C,\tau).
\]
This family captures reward designs in which the evaluator receives both $C$ and $\tau$ and can in principle ignore $\tau$.

\subsection{Predictive Sufficiency of the Core BOW Reward}

The core BOW reward targets predictive sufficiency rather than minimal or lexically compressed representations. Because the core scoring pass receives only $\tau$, reward depends on whether $\tau$ preserves information useful for predicting $Y$. This does not require $\tau$ to be context-free, information-minimal, or lexically disjoint from $C$.

For analysis, we use a log-likelihood-shaped surrogate of the core reward,
\[
\tilde r_{\textsc{bow}}(\tau;Y) := \log P_{J_\phi}(Y \mid \tau),
\]
which is used only in this section.

\begin{theorem}[Predictive sufficiency under a log-reward surrogate]
\label{theorem:sufficiency}
Assume the scorer is approximately calibrated in the sense that
\[
P_{J_\phi}(Y\mid\tau) \approx P(Y\mid\tau).
\]
Then maximizing the expected shaped reward
\[
\max_\theta \ \mathbb{E}_{(C,Y)} \ \mathbb{E}_{\tau \sim \pi_\theta(\cdot\mid C)}
\big[\log P_{J_\phi}(Y\mid \tau)\big]
\]
approximately minimizes the conditional entropy $H(Y\mid \tau)$, equivalently maximizes the mutual information $I(\tau;Y)$.

Moreover, under the Markov structure $\tau \leftarrow C \rightarrow Y$, we have the identity
\[
I(Y;C\mid\tau)=H(Y\mid\tau)-H(Y\mid C),
\]
so minimizing $H(Y\mid\tau)$ also minimizes the remaining predictive information in $C$ beyond $\tau$. This conditional mutual information is the predictive-sufficiency gap: it vanishes exactly when
\[
\begin{aligned}
I(Y;C \mid \tau)=0
\quad \Longleftrightarrow \quad
P(Y \mid C)=P(Y \mid \tau).
\end{aligned}
\]
\end{theorem}

\paragraph{Proof sketch.}
If $P_{J_\phi}(Y\mid\tau)=P(Y\mid\tau)$, then
\[
\begin{aligned}
\mathbb{E}[\log P_{J_\phi}(Y \mid \tau)]
&= \mathbb{E}[\log P(Y \mid \tau)] \\
&= -H(Y \mid \tau).
\end{aligned}
\]
Thus maximizing the expected shaped reward is equivalent to minimizing $H(Y\mid\tau)$ and therefore maximizing
\[
I(\tau;Y)=H(Y)-H(Y\mid\tau).
\]
Next, using $\tau \leftarrow C \rightarrow Y$ (equivalently $I(Y;\tau\mid C)=0$), we have $H(Y\mid C,\tau)=H(Y\mid C)$. Therefore
\[
\begin{aligned}
I(Y;C \mid \tau)
&= H(Y \mid \tau) - H(Y \mid C,\tau) \\
&= H(Y \mid \tau) - H(Y \mid C).
\end{aligned}
\]
Since $H(Y\mid C)$ is fixed by the data distribution, minimizing $H(Y\mid\tau)$ directly minimizes $I(Y;C\mid\tau)$. The achievable minimum depends on the policy class; a zero gap corresponds to predictive sufficiency, $P(Y\mid C)=P(Y\mid\tau)$.

\paragraph{Interpretation.}
Because the core scoring pass receives only $\tau$, improving the core reward requires $\tau$ to preserve information useful for predicting $Y$. Under the assumptions above, the objective encourages predictive sufficiency. This is not a claim that $\tau$ excludes or minimally encodes information from $C$.

\subsection{Two Reward-Bypass Mechanisms}

We consider two ways in which the learning signal for $\tau$ can be bypassed. Under joint context-and-trajectory scoring, the evaluator can predict $Y$ directly from $C$. Under answer-only rewards, the policy can predict $\hat y$ from $C$ without using $\tau$. In either case, improving reward need not make the trajectory predictive.

\begin{theorem}[Reward bypass under joint context-and-trajectory likelihood scoring]
\label{theorem:ctx_likelihood_bypass}
Assume a Bayes-optimal joint scorer, so that
\[
P_{\tilde J_\phi}(Y\mid C,\tau)=P(Y\mid C,\tau).
\]
Under $\tau \leftarrow C \rightarrow Y$, we have
\[
P(Y\mid C,\tau)=P(Y\mid C),
\]
hence
\[
\begin{aligned}
r_{\textsc{ctx}}(C,\tau;Y)
&= \log P_{\tilde J_\phi}(Y \mid C,\tau) \\
&= \log P(Y \mid C).
\end{aligned}
\]
The resulting reward is independent of $\tau$. Consequently, the policy-gradient learning signal for shaping $\pi_\theta(\tau\mid C)$ collapses:
\[
\mathbb{E}_{\tau \sim \pi_\theta(\cdot\mid C)}
\Big[\nabla_\theta \log \pi_\theta(\tau\mid C)\, r_{\textsc{ctx}}(C,\tau;Y)\Big]=0.
\]
\end{theorem}

\paragraph{Proof sketch.}
Because $\tau$ is generated from $C$ and does not affect $Y$, we have $Y \perp \tau \mid C$, and thus $P(Y\mid C,\tau)=P(Y\mid C)$. Therefore the reward is constant with respect to $\tau$ for fixed $(C,Y)$. The score-function identity then gives
\[
\mathbb{E}_{\tau}[\nabla_\theta \log \pi_\theta(\tau\mid C)] = 0,
\]
so multiplying by a constant reward yields zero expected gradient.

\paragraph{Interpretation.}
If the evaluator separately receives $C$, it can recover the next-token distribution without using $\tau$. The trajectory is then no longer necessary for reward, so optimization need not increase $I(\tau;Y)$.

\begin{theorem}[Reward bypass under answer-only rewards]
\label{theorem:ctx_answer_bypass}
A policy $\pi_\theta$ generates a reasoning trajectory $\tau$ and then a predicted next token or word $\hat y$:
\[
\begin{gathered}
(\tau,\hat y)\sim \pi_\theta(\cdot \mid C), \\
\pi_\theta(\tau,\hat y \mid C)
= \pi_\theta(\tau \mid C)\,
  \pi_\theta(\hat y \mid C,\tau).
\end{gathered}
\]
Consider any answer-only reward $g(\hat y,Y)$ that depends on the predicted answer $\hat y$ and target $Y$ but does not directly score $\tau$. This includes BR, where $g(\hat y,Y)=\mathbf{1}\{\hat y=Y\}$.

Define the rollout objective
\[
J(\theta):=\mathbb{E}_{(\tau,\hat y)\sim \pi_\theta(\cdot\mid C)}[g(\hat y,Y)]
\]
under the data distribution of $(C,Y)$ and the policy. Then the policy gradient decomposes as
\[
\begin{aligned}
\nabla_\theta J(\theta)
&=
\mathbb{E}_{\substack{(\tau,\hat y)\sim\\ \pi_\theta(\cdot \mid C)}}
\big[
  g(\hat y,Y)
  \nabla_\theta \log \pi_\theta(\hat y \mid C,\tau)
\big] \\
&\quad+
\mathbb{E}_{\substack{\tau\sim\\ \pi_\theta(\cdot \mid C)}}
\big[
  Q_\theta(C,\tau)
  \nabla_\theta \log \pi_\theta(\tau \mid C)
\big].
\end{aligned}
\]
Here, the post-$\tau$ value is
\[
Q_\theta(C,\tau):=
\mathbb{E}_{\hat y\sim \pi_\theta(\cdot\mid C,\tau)}[g(\hat y,Y)].
\]

Moreover, if the policy reaches (or approximates) a bypass regime in which the answer distribution becomes $\tau$-invariant given $C$, i.e.
\[
\begin{gathered}
\pi_\theta(\hat y \mid C,\tau)
\approx \pi_\theta(\hat y \mid C), \\
\text{for all $\tau$ in the policy support},
\end{gathered}
\]
then $Q_\theta(C,\tau)\approx Q_\theta(C)$ is approximately constant in $\tau$, and the learning signal specifically for shaping the reasoning policy collapses:
\[
\begin{aligned}
&\mathbb{E}_{\tau\sim \pi_\theta(\cdot \mid C)}
\big[
  Q_\theta(C)\nabla_\theta \log \pi_\theta(\tau \mid C)
\big] \\
&\quad =
Q_\theta(C)\,
\mathbb{E}_{\tau\sim \pi_\theta(\cdot \mid C)}
\big[
  \nabla_\theta \log \pi_\theta(\tau \mid C)
\big] \\
&\quad = 0.
\end{aligned}
\]
The answer term
\[
\mathbb{E}\big[g(\hat y,Y)\nabla_\theta \log \pi_\theta(\hat y\mid C,\tau)\big]
\]
can remain nonzero and continue improving prediction accuracy.

Finally, because the objective depends on $\tau$ only through its effect on $\pi_\theta(\hat y\mid C,\tau)$, there exist reward-optimal solutions in which $\hat y$ implements a Bayes-optimal predictor from $C$ (under $g$) while $\tau$ is arbitrary, even independent of $C$. Hence maximizing $J(\theta)$ does not in general force $\tau$ to become an information-sufficient statistic for $Y$.
\end{theorem}

\paragraph{Proof sketch.}
By the policy gradient,
\[
\begin{aligned}
\nabla_\theta J(\theta)
&=
\mathbb{E}_{\substack{(\tau,\hat y)\sim\\ \pi_\theta(\cdot \mid C)}}
\big[
  g(\hat y,Y)
  \nabla_\theta \log \pi_\theta(\tau,\hat y \mid C)
\big].
\end{aligned}
\]
Using the autoregressive factorization
\[
\log \pi_\theta(\tau,\hat y\mid C)
=
\log \pi_\theta(\tau\mid C)+\log \pi_\theta(\hat y\mid C,\tau),
\]
we obtain the stated decomposition. In the bypass regime where $\pi_\theta(\hat y\mid C,\tau)\approx \pi_\theta(\hat y\mid C)$, the post-$\tau$ value becomes approximately constant in $\tau$, so the reasoning-gradient term vanishes by the score-function identity, while the answer-gradient term can remain nonzero and continue improving the answer distribution.

\paragraph{Interpretation.}
BR-style training can improve next-token prediction through the answer term without making $\tau$ self-contained or causally responsible. Once the policy predicts $\hat y$ from $C$ without relying on $\tau$, the gradient shaping $\pi_\theta(\tau\mid C)$ can vanish and the trajectory can become a decorative byproduct. In BOW-Core, the core scorer does not receive $C$ as a separate input, so the reward depends on what is present in $\tau$.

\subsection{Scope of the Bottleneck Claim}

The analysis establishes a structural property: the core scoring pass cannot predict directly from $C$ while ignoring $\tau$. It does not require $\tau$ to be context-free or to avoid restating contextual information. Any context-derived evidence used by the core scorer must first be present in $\tau$. This is the reward-path bottleneck claimed by BOW.

\begin{table*}[t]
  \centering
  \begingroup
  \scriptsize                           
  \setlength{\tabcolsep}{4pt}           
  \renewcommand{\arraystretch}{0.9}     
  \begin{tabularx}{\textwidth}{@{}lXX@{}}
    \toprule
     & \textbf{Original Instance} & \textbf{Transformed Instance} \\
    \midrule
    \textbf{PIQA} &
    \textit{Question}: Choose the most sensible solution given the physical goal: To cream butter and sugar together, you can\par
    \textit{Options}: (a) Place it in a bowl and use a hand warmer (b) Place in a bowl and use a hand mixer &
    \textit{Context}: To cream butter and sugar into a light, fluffy mixture, place them in a bowl and use a hand\par
    \textit{Candidate Next Words}: (a) warmer (b) mixer \\
    \midrule
    \textbf{CSQA} &
    \textit{Question}: Where could you find a shark before it was caught?\par
    \textit{Options}: (a) pool hall (b) tomales bay (c) marine museum (d) business (e) desert &
    \textit{Context}: Before being caught, a shark naturally lives in the\par
    \textit{Candidate Next Words}: (a) hall (b) bay (c) museum (d) business (e) desert \\
    \bottomrule
  \end{tabularx}
  \endgroup
  \caption{Examples of the multiple-choice-to-next-word transformation.}
  \label{tab:transformation_examples}
\end{table*}

\section{Why Soft Partial-Credit Rewards Give Finer Credit Than Binary Hit/Miss Rewards}
\label{appendix:math_soft_vs_binary}

This appendix provides an informal justification for why the \emph{core} BOW reward in Eq.~\ref{eq:core} can yield denser learning signal than a binary exact-match reward such as BR. The discussion concerns BOW-Core; the auxiliary BOW-Reg shaping term in Eq.~\ref{eq:reg} is omitted for clarity.

\subsection{Setup and Notation}

Let $x_{1:t}$ denote the context and let $x^*_{t+1}\in V$ denote the gold next token, where $V$ is the model vocabulary. The policy samples a reasoning trajectory
\[
\tau \sim \pi_\theta(\cdot \mid x_{1:t}).
\]
We compare two reward types:

\paragraph{Binary hit/miss reward.}
The model predicts a next token or word $\hat x_{t+1}$ and receives
\[
R_{\textsc{binary}} = \mathbf{1}\{\hat x_{t+1}=x^*_{t+1}\}\in\{0,1\}.
\]

\paragraph{Soft partial-credit reward.}
A frozen scorer that observes only the reasoning trajectory assigns
\[
R_{\textsc{soft}} = P_{J_\phi}(x^*_{t+1}\mid \tau)\in[0,1].
\]

The corresponding information flow can be summarized informally as:
\[
x_{1:t}\xrightarrow{\pi_\theta}(\tau,\hat x_{t+1})\xrightarrow{\mathbf{1}\{\hat x_{t+1}=x^*_{t+1}\}} R_{\textsc{binary}},
\]
versus
\[
x_{1:t}\xrightarrow{\pi_\theta}\tau\xrightarrow{P_{J_\phi}(x^*_{t+1}\mid \tau)} R_{\textsc{soft}}.
\]

\subsection{Soft Rewards Preserve Partial Credit}

Unlike binary supervision, the soft reward can assign positive and graded credit before the exact gold token is identified. Whenever $P_{J_\phi}(x^*_{t+1}\mid \tau)>0$, the trajectory receives positive reward. A trajectory that narrows the continuation to a plausible class can therefore receive partial credit without identifying the exact token, provided the scorer retains nonzero mass on the gold token. Binary exact-match supervision assigns zero reward unless the model predicts the gold token exactly.

Because the reward equals $P_{J_\phi}(x^*_{t+1}\mid \tau)$, it also preserves the ordering of partially informative trajectories: a trajectory for which the scorer assigns more probability to the gold token. Binary supervision cannot distinguish two trajectories when both miss the exact token.

\subsection{Implication for Exploration}

Soft rewards provide nonzero and graded feedback on a broader set of rollouts, which can reduce dependence on rare exact-match successes. This is an argument about the granularity of credit assignment, not a guarantee that soft rewards dominate binary rewards in every optimization regime.

\begin{figure*}[!t]
\centering
\begin{minipage}[t]{\textwidth}
\centering
\begin{tcolorbox}[
    top=0.5mm,
    bottom=0.5mm,
    left=0.5mm,
    right=0.5mm
]
\ttfamily\scriptsize
The next word set will likely contain nouns that represent beverages, as the runner is at an aid station and is likely to grab a drink to rehydrate. \\ \\
The set of possible next words can be described as: \\ - Water \\ - Sports drink \\ - Juice  \\ - Electrolyte solution \\ \\
These are the most common types of beverages found at aid stations during marathons, designed to help runners rehydrate and replenish electrolytes lost through sweat.
\end{tcolorbox}
\end{minipage}

\vspace{0.2cm} 

\begin{minipage}[t]{\textwidth}
\centering
\resizebox{0.6\textwidth}{!}{%
\begin{tabular}{cccccccc}
\toprule
 & & \textbf{\textit{water}} & \textbf{\textit{sports}} & \textbf{\textit{electro}} & \textbf{\textit{juice}} & \textbf{\textit{drink}} & \textbf{\textit{energy}} \\
\midrule
\multirow{2}{*}[-0.3em]{\textbf{LLaMA3.1-8B-Instruct}} & \textbf{Rank} & 1 & 5 & 7 & 10 & 11 & 16 \\
\cmidrule{2-8}
 & \textbf{Prob.} & 0.7575 & 0.0228 & 0.0074 & 0.0024 & 0.0011 & 0.0007 \\
\midrule
\multirow{2}{*}[-0.3em]{\textbf{Qwen2.5-7B-Instruct}} & \textbf{Rank} & 1 & 2 & N/A & 13 & N/A & N/A \\
\cmidrule{2-8}
 & \textbf{Prob.} & 0.9995 & 0.0004 & N/A & 0.0000 & N/A & N/A \\
\bottomrule
\end{tabular}%
}
\end{minipage}
\caption{This example shows that several expected next tokens appear near the top of $P_{J_\phi}(\cdot \mid \tau)$. Given the context \textit{The marathon runner felt dizzy from dehydration, so at the aid station she grabbed a bottle of}, the trajectory shown above supports beverage-related continuations. Both scorers rank \textit{water} first; Qwen2.5-7B-Instruct concentrates nearly all probability mass on it (0.9995), whereas LLaMA3.1-8B-Instruct remains less concentrated (0.7575) and assigns non-negligible mass to additional plausible candidates.}
\label{fig:scorer_justification}
\end{figure*}

\section{More Experimental Details}

\subsection{Training Data Filtering}
\label{appendix:training_data_filtering}
We filter the training data to remove context--next-word pairs whose next token can be derived without meaningful reasoning over the context. Specifically, we discard tokens that are:
(i) purely functional (e.g., determiners, punctuation),
(ii) syntactically or semantically deterministic based on surface cues, or
(iii) explainable without invoking latent knowledge or contextual abstraction. This filtering strategy is inspired by selective language modeling methods such as RHO-1 \citep{lin2024not}, which demonstrates that focusing training on informative or "reasoning-heavy" tokens improves learning efficiency and model generalization.
We use \texttt{gpt-4.1-mini-2025-04-14}\footnote{\url{https://openai.com/index/gpt-4-1/}} to evaluate each context--next-word pair using these criteria. Fig.~\ref{fig:training-data-filtering-prompt} shows the filtering prompt. We retain only pairs whose next word requires non-trivial reasoning.

\subsection{Targeted Ambiguity Evaluation Details}
\label{appendix:ambiguity_details}

For SharedRef, we evaluate on the English SharedRef split under the Normal prompt. This setting preserves referential ambiguity while avoiding the additional simplified-language constraint. We report the benchmark's official judge-based Correctness metric.
For HoWN-Simple, we evaluate English homonyms under the Simple prompt, which asks for definitions in simple language without the Multi-Sense-Aware hint. We report Covered and Collapse. Covered is the percentage of coarse-grained WordNet senses covered by the response. Collapse is the rate of responses that provide only one definition without helpful sense awareness. We omit Complete, FKGL, and other HoWN diagnostics from the main table because Covered measures actual sense coverage, while Collapse directly measures the premature-commitment failure mode targeted by BOW.

\subsection{NWP Format Transformation}
\label{appendix:data_transformation}
We use \texttt{gpt-4.5-preview-2025-02-27}\footnote{\url{https://openai.com/index/introducing-gpt-4-5/}} to jointly transform each multiple-choice QA instance into a context and a set of single-word candidate continuations, with one gold candidate. Each candidate must complete the context naturally, and the transformed task must preserve the original knowledge, reasoning process, and approximate difficulty. We then use GPT-4.5 to filter out invalid transformations. Figs.~\ref{fig:eval_data_transform_prompt} and~\ref{fig:eval_data_transform_validation_prompt} show the transformation and validation prompts.

\subsection{More Optimization Details}
\label{appendix:more_optim_details}
We conduct RL training on four NVIDIA H200 GPUs using the VeRL\footnote{\url{https://github.com/volcengine/verl}} \citep{10.1145/3689031.3696075} repository.
We use the AdamW \citep{loshchilov2018decoupled} optimizer with an initial learning rate of $1\times10^{-6}$, ($\beta_1,\beta_2$) = ($0.9, 0.999$), and a weight decay of $1\times10^{-2}$.

For representative Qwen2.5-7B-Instruct runs, one epoch on four NVIDIA H200 GPUs took approximately 6 wall-clock hours for BR, 9 for BOW-Core, and 12 for BOW-Reg, corresponding to about 24, 36, and 48 H200 GPU-hours, respectively. BOW-Core and BOW-Reg therefore used roughly $1.5\times$ and $2.0\times$ the GPU-hours of BR. The additional cost comes from frozen-scorer inference over the sampled trajectories; BOW-Reg also computes the context-based reference distribution used by the breadth regularizer.

\begin{table*}[!t]
\centering
\resizebox{1\linewidth}{!}{%
\setlength{\tabcolsep}{6pt}
\small
\begin{tabular}{
l
*{11}{S[table-format=+2.2, table-space-text-post=\downarrowcell]}
}
\toprule
& \multicolumn{1}{c}{\textbf{CSQA}}
& \multicolumn{1}{c}{\textbf{PIQA}}
& \multicolumn{1}{c}{\textbf{SQA}}
& \multicolumn{1}{c}{\textbf{TQA}}
& \multicolumn{1}{c}{\textbf{ARC-c}}
& \multicolumn{1}{c}{\textbf{WG}}
& \multicolumn{1}{c}{\textbf{BBH}}
& \multicolumn{1}{c}{\textbf{MMLU}}
& \multicolumn{1}{c}{\textbf{MMLU-p}}
& \multicolumn{1}{c}{\textbf{GPQA}}
& \multicolumn{1}{c}{\textbf{AVG.}} \\

\midrule \\[-13pt]
\multicolumn{12}{c}{\footnotesize\texttt{\textit{Qwen2.5-7B-Instruct}}} \\[-2.5pt]
\midrule

BOW-Reg
& 81.90 & 88.68 & 63.19 & 66.83 & 92.32 & 70.96 & 77.21 & 76.51 & 57.51 & 33.26 & 70.84 \\

\hspace{0.5em}\textit{w/ random filt.}
& 0.00
& -0.84\downarrowcell
& -1.02\downarrowcell
& -1.22\downarrowcell
& +0.60\uparrowcell
& -0.24\downarrowcell
& +0.21\uparrowcell
& -0.34\downarrowcell
& +0.27\uparrowcell
& 0.00
& -0.26\downarrowcell \\

\midrule \\[-13pt]
\multicolumn{12}{c}{\footnotesize\texttt{\textit{LLaMA3.1-8B-Instruct}}} \\[-2.5pt]
\midrule

BOW-Reg
& 77.97 & 86.67 & 67.28 & 60.22 & 89.16 & 59.75 & 74.56 & 73.59 & 52.71 & 32.14 & 67.41 \\

\hspace{0.5em}\textit{w/ random filt.}
& +0.41\uparrowcell
& -0.87\downarrowcell
& -0.20\downarrowcell
& -0.86\downarrowcell
& +0.35\uparrowcell
& +0.94\uparrowcell
& -3.01\downarrowcell
& -1.41\downarrowcell
& -3.68\downarrowcell
& -2.45\downarrowcell
& -1.08\downarrowcell \\

\bottomrule
\end{tabular}
}
\caption{Effect of reasoning-centric training-data filtering. BOW-Reg rows report absolute scores; random-filtering rows report changes relative to BOW-Reg.}
\label{tab:random_filter_ablation}
\end{table*}

\subsection{More Baseline Details}
\label{appendix:more_baseline_details}

For fair comparison, all trainable baselines use the same filtered supervision positions as BOW.

\paragraph{ToW.}
For ToW annotations, we use \texttt{gpt-4o-2024-11-20}.\footnote{\url{https://openai.com/index/hello-gpt-4o/}} Following \citet{xu-etal-2025-tow}, we provide the context, three in-context examples, and the gold next word, then ask the model to generate a thought of about 20 words. We annotate the same supervision positions used for RL rather than applying the original three-stage generation, filtering, and shortening pipeline. As in prior work, the model may state that the next word is unpredictable from the context.

\paragraph{BR.}
BR is designed to match BOW as closely as possible while remaining answer-centric. The policy is prompted to reason about the most likely next word and then output a single predicted next word wrapped in \texttt{\textbackslash boxed\{\}}; the exact prompt is shown in Fig.~\ref{fig:no_judge_prompt}. Reward is 1 if the first token of the boxed word matches the first token of the gold next word, and 0 otherwise.

\paragraph{IGR.}
Because the original RLP implementation is not publicly available, we implement a simplified information-gain baseline. We use the initial policy model as a frozen reference scorer and compute reward as the difference between the log-evidence assigned to the gold next token with reasoning and the corresponding baseline log-evidence without reasoning.

\paragraph{Intrinsic NWP evaluation details.}
BR directly emits a boxed next word rather than a candidate-scored distribution. For a fair comparison in intrinsic NWP evaluation, we remove the boxed answer and score the benchmark candidate continuations from the preceding reasoning prefix.

\section{Human Evaluation Details and Representative Examples}
\label{appendix:more_human_study_details}

This appendix describes the curation and annotation of the 150 human-evaluation contexts, followed by three representative examples of premature collapse from a broader plausible next-word space to a single favored continuation.

\subsection{Context Curation}

We curate 150 contexts in which several immediate next words are superficially plausible but contextual reasoning narrows the continuation space. GPT-4.5 selects contexts with two properties: (i) several continuations are plausible from local statistical, grammatical, or commonsense cues, and (ii) broader contextual reasoning narrows the space to one or a few especially plausible candidates. The selected next words include nouns, verbs, and adjectives.

\subsection{Annotation Protocol}

We compare four Qwen-based models: the vanilla instruction model, BR, BOW-Core, and BOW-Reg. For each curated context, we sample one next-word reasoning trajectory from each model. Annotators are shown the original context together with four anonymized and randomly ordered trajectories. Each set of trajectories is evaluated along two dimensions:

\begin{enumerate}
    \item \textbf{Comprehensive next-word reasoning.} Which trajectory best reasons over the plausible next-word space in a self-contained, neutral, and comprehensive way, using clues expressed or implied by the context without collapsing too early to a narrow answer?
    \item \textbf{Human-expected next-word description.} Which trajectory's next-word description best matches the annotator's own expected continuation?
\end{enumerate}

Ties are allowed when no clear winner is apparent, so the reported preference rates do not sum to 100.

\subsection{Interpreting the Annotation Dimensions}

The first dimension directly measures the trajectory form targeted by BOW. The second is diagnostic: it measures whether the trajectory states the continuation an annotator expects most directly. A trajectory may score highly on the second dimension by committing early, while another may score more highly on comprehensiveness by preserving a broader candidate set.

\subsection{Representative Examples}

Tabs.~\ref{tab:example_thunder}, \ref{tab:example_high_stake_1}, and \ref{tab:example_high_stake_2} illustrate the qualitative differences behind the quantitative results. In these tables, the original context is shown alongside the trajectories only for the reader. The policy conditions on that context when generating each self-contained trajectory, so a trajectory may restate parts of it; the core scoring pass receives only the trajectory.

\paragraph{Ordinary ambiguous context.}
For the context \textit{The thunderstorm was getting closer, so I rolled up the \ldots}, Vanilla and BR commit quickly to \textit{window}. BOW-Core names a somewhat broader but still explicit set, while BOW-Reg remains less committed and describes a wider range of weather- and roll-up-related possibilities. Its broader wording also helps explain its lower preference on expected next-word description.

\paragraph{Medical example.}
In Tab.~\ref{tab:example_high_stake_1}, BR and BOW-Core move quickly toward \textit{antibiotics} or \textit{antibiotic}. BOW-Reg stays at a broader treatment-related level rather than immediately committing to one continuation.

\paragraph{Finance example.}
In Tab.~\ref{tab:example_high_stake_2}, BR and BOW-Core commit to \textit{transfers}. BOW-Reg discusses broader concepts around control transfer instead of treating that continuation as uniquely licensed.

Across the three examples, BOW-Core is more willing to state a likely continuation directly, while BOW-Reg more strongly resists early commitment.

\section{Effect of Training Data Filtering}
\label{sec:data_filtering}

Tab.~\ref{tab:random_filter_ablation} compares the reasoning-centric filter with random filtering that retains the same number of supervision positions. Random filtering lowers average performance by 0.26 points on Qwen2.5-7B-Instruct and 1.08 points on LLaMA3.1-8B-Instruct. The filter is not the main source of BOW's gains; it concentrates updates on contexts where next-word reasoning is needed.

\section{Use of AI Assistants}
We used AI assistants to polish language and presentation throughout the manuscript based on author-provided research ideas, experimental design, analyses, and results. AI assistants were not used to originate the research questions or generate plots; the authors reviewed and take responsibility for all content.

As part of the experimental pipeline, we used \texttt{gpt-4.1-mini-2025-04-14} for training-data filtering, \texttt{gpt-4.5-preview-2025-02-27} for evaluation-data transformation and human-evaluation context curation, and \texttt{gpt-4o-2024-11-20} for ToW annotations. We also used LLM-based judging for SharedRef and HoWN-Simple, as described above.

\begin{figure*}[!t]
\centering
\begin{tcolorbox}[
    title=Policy rollout prompt,
    top=1mm,
    bottom=1mm,
    left=1mm,
    right=1mm
]
\ttfamily\scriptsize
The next word prediction problem of language modeling generally involves first searching for a comprehensive but small enough set of possible next words, and then estimating each possible word’s probability within this set. Given a specific context, your job is to output a reasoning process that generally follows this illustration.\\ \\
Here is one important rule you MUST follow. Since different candidate sets are different in size, instead of enumerating all possible words, you must be smart and innovative enough to describe the word set in concise and unbiased natural language, without missing any possibilities. This is particularly important when the smallest candidate set you can find is still not small or concrete enough. For example, given the context ``I [next-word]'', the best you can do is to tell me that the next word set will likely contain verbs, and those verbs should be actions performed by humans rather than other verbs. You should not give a concrete list of words since it is impossible to enumerate all the verbs humans can perform.\\ \\
However, in other contexts where information is rich, you should perform more in-depth reasoning. For example, given the context ``I like fruit, so as lunch I had two [next-word]'', you should reason that the next word set will likely contain nouns of fruit given the logical clue that ``I like fruit'' and the quantitative adjective ``two'' in the context. Furthermore, you should reason that this set is more likely to contain larger fruits to fill me up for lunch instead of small fruit. In other scenarios, such as ``1+1=[next-word]'', where the next word set can be small enough to be listed out, you should actually list out all possible words in the set you find. \\ \\
Now, given the below context, please reason on possible next words following all the rules described above. Make your output self-contained without needing to reference the given context.\\
Context: \\
\{context\}
\end{tcolorbox}
\caption{Policy rollout prompt.}
\label{fig:policy_prompt}
\end{figure*}

\begin{figure*}[!t]
\centering
\begin{tcolorbox}[
    title=Binary-reward rollout prompt,
    top=1mm,
    bottom=1mm,
    left=1mm,
    right=1mm
]
\ttfamily\scriptsize
The next word prediction problem of language modeling generally involves first searching for a comprehensive but small enough set of possible next words, and then estimating each possible word's probability within this set to pick the most likely next word. Given a specific context, your job is to output a reasoning process that generally follows this illustration to predict the most likely next word.\\ \\
Here is one important rule you MUST follow. Since different candidate sets are different in size, instead of enumerating all possible words, you must be smart and innovative enough to describe the word set in concise and unbiased natural language, without missing any possibilities. This is particularly important when the smallest candidate set you can find is still not small or concrete enough. For example, given the context "I like fruit, so as lunch I had two [next-word]", you should reason that the next word set will likely contain nouns of fruit given the logical clue that "I like fruit" and the quantitative adjective "two" in the context. Furthermore, you should reason that this set is more likely to contain larger fruits to fill me up for lunch instead of small fruit. Then you should choose the most likely next word based on the set description. \\ \\
Now, given the below context, please follow the rule described above to first reason and then wrap your predicted most likely next word within \textbackslash boxed\{\}. Make your output self-contained without needing to reference the given context. \\
Context: \\
\{context\}
\end{tcolorbox}
\caption{Binary-reward rollout prompt.}
\label{fig:no_judge_prompt}
\end{figure*}

\begin{figure*}[!t]
\centering
\begin{tcolorbox}[
    title=Frozen-scorer prompt,
    top=1mm,
    bottom=1mm,
    left=1mm,
    right=1mm
]
\ttfamily\scriptsize
Task Instruction: Given a thought for the next word of a certain context, you need to predict this next word. \\ \\
Now please give me your prediction of the next word based on the following thought: \\ \\
\{thought\} \\ \\
Next Word:
\end{tcolorbox}
\caption{Frozen-scorer prompt.}
\label{fig:judge_prompt}
\end{figure*}

\begin{figure*}[!t]
\centering
\begin{tcolorbox}[
    title={Training-data filtering prompt},
    top=1mm,
    bottom=1mm,
    left=1mm,
    right=1mm
]
\ttfamily\scriptsize
Given a context and its completion, you need to decide if the context’s immediate next word, which is the first word in the completion, requires non-trivial reasoning to derive. Below are two examples:\\ \\
<Context>\\
During her last visit, my mom gave me a pair of socks and a swimsuit. ......(words omitted here). This winter, to stay warm, I put on a pair of\\
</Context>\\
<Completion>\\
socks that my mom gave me last time.\\
</Completion>\\
In this example, the immediate next word to the context is "socks", and it requires non-trivial reasoning to derive. It is common sense to know that when people feel cold, they will put on warm clothes to keep warm. Some possible next words to the context could be “jeans”, “socks”, or “gloves”, etc. However, from the context, we know that mom gave me a pair of socks last time, so it is more likely for me to put on a pair of socks instead of gloves or jeans to keep warm, since it is known that I have this pair of socks.\\ \\
<Context>\\
During her last visit, my mom gave me a pair\\
</Context>\\
<Completion>\\
of socks and a swimsuit.\\
</Completion>\\
In this example, the immediate next word to the context is “of”, and it does not require reasoning to derive. Given “a pair”, English grammar already tells us the next word is highly likely to be “of” to complete the phrase “a pair of”.\\ \\
The two examples above are simplified examples to illustrate your goal; the context you are about to see is longer and more complex. Now, I am providing you with a context and its completion. Please decide if the immediate next word to the context, which is the first word in the completion, requires non-trivial or complex reasoning to derive:\\
<Context>\\
\{context\}\\
</Context>\\
<Completion>\\
\{completion\}\\
</Completion>\\ \\
Note that you must respond in the format of a JSON object with two keys. The first key is named “requires\_reasoning” with Boolean type to decide if reasoning is required. True means required, and false means not required. The second key is called “explanation” with String type, and it should record your explanation of why or why not non-trivial reasoning is needed.
\end{tcolorbox}
\caption{Training-data filtering prompt.}
\label{fig:training-data-filtering-prompt}
\end{figure*}

\begin{figure*}[!t]
\centering
\begin{tcolorbox}[
    title=MCQA inference prompt,
    top=1mm,
    bottom=1mm,
    left=1mm,
    right=1mm
]
\ttfamily\scriptsize
Question: \{question\} \\
Options: \{options\} \\ \\
You should ONLY choose the letters from the options and answer in the format of 'So the answer is X', where X is the option letters A, B, C, etc. Please think step by step.
\end{tcolorbox}
\caption{MCQA inference prompt.}
\label{fig:mcqa_inference_prompt}
\end{figure*}

\begin{figure*}[!t]
\centering
\begin{tcolorbox}[
    title=Evaluation transformation prompt,
    top=1mm,
    bottom=1mm,
    left=1mm,
    right=1mm
]
\ttfamily\scriptsize
\# Task:
Given a question, its candidate answers, and the correct answer, your task is to transform them into a final word prediction problem. \\ \\
\# Transformation rules:\\
1. The transformation output should contain a transformed context, a set of candidate final words (must be WORD instead of phrase), and the gold final word, which is much more likely to conclude the transformed context than other candidate final words.\\
2. Concatenating each candidate final word to the transformed context must be able to form a COMPLETE sentence. This complete sentence must be NATURAL and FLUENT.\\
3. The transformed final word prediction problem must maintain the SAME CONTEXT from the original question, so that it can evaluate the SAME or roughly the same KNOWLEDGE and REASONING process from the original question at the SAME DIFFICULTY level.\\

\# Output format requirements:\\
You must output in JSON format with three keys. The first key is named ``transformed\_context'' with the data type of string to store the transformed context. The second key is named ``final\_word\_candidates'' with the data type of a list of strings to store the transformed candidate final words. The third key is named `` gold\_final\_word\_index'' with the data type of integer to store the gold final word index in the ``final\_word\_candidates'' list. \\ \\
Now, you are given the following question, its candidate answers, and the gold answer. Please follow the above transformation rules, example transformations, and output format requirements to perform the transformation.\\ \\
<Original Question>\\
\{question\}\\
</Original Question>\\
<Original Candidate Options>\\
\{options\}\\
</Original Candidate Options> \\
<Original Ground Truth>\\
\{ground\_truth\}\\
</Original Ground Truth>
\end{tcolorbox}
\caption{Evaluation transformation prompt.}
\label{fig:eval_data_transform_prompt}
\end{figure*}

\begin{figure*}[!t]
\centering
\begin{tcolorbox}[
    title=Transformation validation prompt,
    top=1mm,
    bottom=1mm,
    left=1mm,
    right=1mm
]
\ttfamily\scriptsize
\# Task:
Last word prediction is the task of predicting the last word of a given context. Given an instance of last word prediction, there is a context, a set of candidate last words for the context, and the gold last word. Your task is to verify the quality of this instance based on the following verification dimensions: \\ \\
\# Verification rules:\\
1. You need to ensure that the gold last word is more plausible than other candidates without any ambiguity.\\
2. All the candidate's last words must be able to be directly concatenated to the context to form a semantically complete sentence.\\
3. The semantically complete sentence from \#2 must be semantically fluent and as natural as possible, just like how we humans would talk in real life. Minor grammatical errors are acceptable.\\

\# Output format rules:\\
You must output in JSON format with two keys. The first key is named ``is\_valid'' with the data type of boolean to store whether the given instance is valid based on the verification rules. The second key is called ``rationale'' with the data type of string to store the rationale of why or why not the instance is valid.\\ \\
Now, given a last word prediction instance and the above verification rules, the instance is not valid if a single rule is violated. Please verify this instance:\\ \\
<Context>\\
\{context\}\\
</Context>\\
<Candidate last words>\\
\{candidate\_last\_words\}\\
</Candidate last words> \\
<Gold last word>\\
\{gold\_last\_word\}\\
</Gold last word>
\end{tcolorbox}
\caption{Transformation validation prompt.}
\label{fig:eval_data_transform_validation_prompt}
\end{figure*}

\begin{table*}[!t]
\small
\setlength{\tabcolsep}{8pt}

\resizebox{1\textwidth}{!}{

\begin{tabular}{@{}ll@{}}
\toprule
\textbf{Context} & \textbf{The thunderstorm was getting closer, so I rolled up the} \\ \midrule
\textbf{Vanilla} &
\begin{minipage}{\linewidth} \sffamily
Given the context "The thunderstorm was getting closer, so I rolled up the," the next word set will likely contain nouns that refer to parts of a window or curtains, as the action of rolling up suggests something that can be rolled. The most probable nouns in this context are "window" and "curtains," as these are the most common items that one would roll up to protect oneself from a thunderstorm. \\ \\
Therefore, the concise and unbiased set of possible next words can be described as "window" or "curtains."
\end{minipage} \\ \midrule
\textbf{BR} &
\begin{minipage}{\linewidth} \sffamily
Reasoning: To predict the most likely next word, we need to consider the context provided: "The thunderstorm was getting closer, so I rolled up the". Given that a thunderstorm is approaching, the action of rolling something up is likely in response to this weather event. The most probable object to be rolled up in this context would be a window or a door, as these are common actions taken to protect oneself from a storm. \\ \\
Thus, the most likely next word set would be: "window" or "door". \\ \\
\textbackslash boxed\{window\}
\end{minipage} \\ \midrule
\textbf{BOW-Core} &
\begin{minipage}{\linewidth} \sffamily
Given the context "The thunderstorm was getting closer, so I rolled up the [next-word]", the next word in the sentence could be "window," "car window," "shade," "curtain", or "blanket," among others. The most common and logical next word in this context would be "window" or "car window," as it continues the phrase "rolled up the" in a way that makes sense in the context of a thunderstorm. The phrase "rolled up the window" is a common expression used when someone closes a car window by rolling it up. Other possibilities like "shade" or "curtain" are also plausible in certain contexts, such as rolling up a window shade or a curtain, but "window" is the most direct and commonly used option. \\ \\
In concise and unbiased natural language, the possible next words for "The thunderstorm was getting closer, so I rolled up the [next-word]" can be "window," "car window," "shade," "curtain", or similar items that can be rolled up, such as a blanket or a curtain. However, "window" is the most likely and commonly used next word in this context.
\end{minipage} \\ \midrule
\textbf{BOW-Reg} &
\begin{minipage}{\linewidth} \sffamily
Given the context "The thunderstorm was getting closer, so I rolled up the," the next word set can be reasoned as follows: \\
1. **Continuation of the sentence**: The sentence is describing an action related to a thunderstorm, so the next words are likely to continue the narrative or provide more details about the situation. \\
2. **Context clues**: The phrase "The thunderstorm was getting closer, so I rolled up the" suggests that the speaker is taking some action in response to the approaching thunderstorm. The word "rolled up" is incomplete and could be referring to something that can be rolled up, such as a window, a curtain, or a tent. \\
3. **Topic relevance**: Since the context mentions a thunderstorm, the next words are likely to be related to weather-related actions or objects, or they might provide more details about the setting or the speaker's actions. \\ \\
Based on these observations, the next word set could likely contain: \\
- Nouns or noun phrases related to weather or outdoor activities (e.g., window, curtain, tent, car window) \\
- Nouns or noun phrases related to the action of rolling up (e.g., window, curtain, tent) \\
- Nouns or noun phrases related to the setting or the speaker's environment (e.g., car, door, side) \\
- Phrases or sentences that continue the narrative or provide more details about the thunderstorm or the speaker's actions \\ \\
Therefore, a concise and unbiased description of the next word set could be:
"The next word set will likely contain nouns or noun phrases related to weather, outdoor activities, the action of rolling up, the setting or the speaker's environment, or phrases that continue the narrative or provide more details about the thunderstorm or the speaker's actions."
\end{minipage}\\
\bottomrule
\end{tabular}
}
\caption{A representative example from the 150 contexts curated in Sec.~\ref{sec:human_results}. The backbone model is Qwen2.5-7B-Instruct.}
\label{tab:example_thunder}
\end{table*}

\begin{table*}[!t]
\footnotesize
\setlength{\tabcolsep}{8pt}
\resizebox{1\textwidth}{!}{%
\begin{tabular}{@{}ll@{}}
\toprule
\textbf{Context} & \textbf{In the ED with fever, hypotension, and suspected sepsis; after blood cultures, initiate broad-spectrum} \\ \midrule
\textbf{Vanilla} &
\begin{minipage}{\linewidth} \sffamily
Given the context "In the ED with fever, hypotension, and suspected sepsis; after blood cultures, initiate broad-spectrum", the next word set will likely contain nouns or adjectives that describe antibiotics or antimicrobial agents, as these are typically used in such clinical scenarios to treat suspected infections, especially in cases of sepsis. \\ \\
The reasoning is as follows: \\
1. The context mentions "fever, hypotension, and suspected sepsis," which indicate a severe infection requiring urgent medical intervention. \\ \\
2. "After blood cultures" suggests that the healthcare provider is awaiting the results of the cultures before proceeding with treatment. \\ \\
3. "Initiate broad-spectrum" indicates that the next word should describe an antibiotic or antimicrobial agent that covers a wide range of potential pathogens. \\ \\
Therefore, the next word set will likely include terms such as "antibiotics," "antimicrobials," or specific names of broad-spectrum antibiotics like "ciprofloxacin," "meropenem," "vancomycin," etc. However, since listing all possible broad-spectrum antibiotics would be impractical, we can describe the set concisely as "broad-spectrum antibiotics or antimicrobial agents."
\end{minipage} \\ \midrule
\textbf{BR} &
\begin{minipage}{\linewidth} \sffamily
Reasoning: Given the context "In the ED with fever, hypotension, and suspected sepsis; after blood cultures, initiate broad-spectrum", the most likely next word is "antibiotics". This is because in a medical setting where a patient presents with fever, hypotension, and suspected sepsis, the standard protocol is to start broad-spectrum antibiotics immediately after obtaining blood cultures. The phrase "broad-spectrum" indicates that the next word should be a noun that can be used in a medical context to treat a wide range of bacterial infections, and "antibiotics" is the most appropriate and commonly used term in such a scenario. \\ \\
\textbackslash boxed\{antibiotics\}
\end{minipage} \\ \midrule
\textbf{BOW-Core} &
\begin{minipage}{\linewidth} \sffamily
Given the context "In the ED with fever, hypotension, and suspected sepsis; after blood cultures, initiate broad-spectrum [next-word]", the next word set is likely to be a noun that describes a type of antibiotic or treatment. The phrase "broad-spectrum" indicates that the next word should be a noun that refers to a medication capable of treating a wide range of bacterial infections, which is common in cases of suspected sepsis. \\ \\
The most appropriate and commonly used next word in this context would be "antibiotic." This noun fits the logical flow of the sentence, as it is the typical next step in managing a patient with suspected sepsis after obtaining blood cultures. \\ \\
While "antibiotic" is the most direct and common choice, the set of possible next words could also include other medical terms that are relevant in this context, such as "therapy," "antimicrobial," or "treatment." However, these are less specific and less commonly used in the immediate context of initiating a broad-spectrum approach in the emergency department. \\ \\
Therefore, the most reasonable and concise next word set for this context is "antibiotic," as it is the most direct and commonly used term in medical practice for initiating broad-spectrum treatment in cases of suspected sepsis.
\end{minipage} \\ \midrule
\textbf{BOW-Reg} &
\begin{minipage}{\linewidth} \sffamily
Given the context "In the ED with fever, hypotension, and suspected sepsis; after blood cultures, initiate broad-spectrum", the next word set can be reasoned as follows: \\
1. **Medical Context**: The context is clearly in a medical setting, specifically in an Emergency Department (ED) dealing with a patient who has fever, hypotension, and suspected sepsis. This sets the tone for the next words to be related to medical treatment or procedures. \\ \\
2. **Action Following Blood Cultures**: The phrase "after blood cultures" indicates that the next words should describe an action or treatment that follows the completion of blood cultures. This could include starting a specific type of treatment, ordering further tests, or documenting the patient's condition. \\ \\
3. **Broad-Spectrum Treatment**: The phrase "initiate broad-spectrum" suggests that the next words will likely describe a type of medication or treatment that is broad-spectrum in nature, meaning it is effective against a wide range of pathogens. \\ \\
Based on these points, the next word set will likely contain: \\
- **Medical treatments or medications**: Such as antibiotics, antivirals, or other broad-spectrum drugs. \\ \\
- **Procedures or actions**: Such as starting an IV, ordering additional tests, or documenting the patient's condition. \\ \\
- **Medical terms**: Such as the names of specific conditions, treatments, or medications. \\ \\
Therefore, a concise and unbiased description of the next word set could be:
"The next word set will likely contain medical treatments or medications, procedures or actions following the completion of blood cultures, and medical terms related to the patient's condition or the treatment being initiated."
\end{minipage}\\
\bottomrule
\end{tabular}
}
\caption{A representative medical example using Qwen2.5-7B-Instruct as the backbone model.}
\label{tab:example_high_stake_1}
\end{table*}

\begin{table*}[!t]
\small
\setlength{\tabcolsep}{8pt}
\resizebox{1\textwidth}{!}{%
\begin{tabular}{@{}ll@{}}
\toprule
\textbf{Context} & \textbf{Under IFRS 15, revenue is recognized when control} \\ \midrule
\textbf{Vanilla} &
\begin{minipage}{\linewidth} \sffamily
Given the context "Under IFRS 15, revenue is recognized when control", the next word set will likely contain nouns or adjectives that complete the phrase "when control [next-word]". The most probable next words would be related to the transfer of risk and reward, which are key elements in determining when revenue is recognized under IFRS 15. \\ \\
Therefore, the next word set can be described as containing terms such as "transfers", "is transferred", "has been transferred", "has transferred", or "will transfer". These terms capture the idea of the transfer of control, which is central to the revenue recognition criteria under IFRS 15. \\ \\
To summarize, the next word set is best described as: "terms indicating the transfer of control, such as 'transfers', 'is transferred', 'has been transferred', 'has transferred', or 'will transfer'."
\end{minipage} \\ \midrule
\textbf{BR} &
\begin{minipage}{\linewidth} \sffamily
Reasoning: Given the context "Under IFRS 15, revenue is recognized when control", the most likely next word is "transfers". This is because IFRS 15, the International Financial Reporting Standard for revenue from contracts with customers, specifies that revenue should be recognized when control of the promised goods or services is transferred to the customer. The word "transfers" directly aligns with the concept of control transfer in the context of revenue recognition under IFRS 15. \\ \\
\textbackslash boxed\{transfers\}
\end{minipage} \\ \midrule
\textbf{BOW-Core} &
\begin{minipage}{\linewidth} \sffamily
Given the context "Under IFRS 15, revenue is recognized when control [next-word]", the next word in the sentence is likely to be a noun that completes the phrase "when control [next-word]". The most logical and common completion for this phrase in the context of accounting and financial reporting standards would be "transfers" to "the customer." It fits grammatically and semantically, as it is used in the context of when revenue is recognized according to IFRS 15. \\ \\
The candidate set for the next word is small and concrete enough to be listed out, and the most appropriate and commonly used word in this context is "transfers." Therefore, the next word in the sentence is most likely "transfers." \\ \\
Output: transfers
\end{minipage} \\ \midrule
\textbf{BOW-Reg} &
\begin{minipage}{\linewidth} \sffamily
Given the context "Under IFRS 15, revenue is recognized when control", the next word set will likely contain nouns, verbs, or phrases that logically follow the discussion of revenue recognition under IFRS 15. The phrase "Under IFRS 15, revenue is recognized when control" suggests that the following words will provide more details about the criteria for recognizing revenue or possibly introduce a new aspect of the standard. \\ \\
To reason further, IFRS 15 (International Financial Reporting Standards 15) is specifically about revenue recognition. Therefore, the next word set will likely include: \\
1. **Nouns or phrases related to revenue recognition criteria**: Such as "transferred" (indicating the transfer of control), "to a customer" (specifying the recipient of the revenue), or "to the customer" (further clarifying the point of transfer). \\ \\
2. **Verbs or phrases that continue the explanation**: Such as "is transferred" (completing the sentence about revenue recognition), "to the customer" (further detailing the process), or "is determined" (introducing a new aspect of the standard). \\ \\
3. **Phrases that introduce a new aspect of the standard**: Such as "of the good or service" (further specifying what is being recognized as revenue), or "is transferred" (continuing the explanation of the criteria). \\ \\
Thus, a concise and unbiased description of the next word set could be: \\
"The next word set will likely contain nouns, verbs, or phrases related to the criteria for recognizing revenue under IFRS 15, specifically focusing on the transfer of control, the recipient of the revenue, and the process of revenue recognition."
\end{minipage}\\
\bottomrule
\end{tabular}
}
\caption{A representative finance example using Qwen2.5-7B-Instruct as the backbone model.}
\label{tab:example_high_stake_2}
\end{table*}

\end{document}